\documentclass{bmvc2k}

\usepackage{amssymb}
\usepackage{amsmath, bm}
\usepackage{amsthm}
\usepackage{booktabs}
\usepackage{algorithm}
\usepackage{algorithmic}
\usepackage{mathtools}

\title{Latent Inversion with Timestep-aware Sampling for Training-free Non-rigid Editing\footnote{This manuscript has been submitted to Pattern Recognition Letters.}}

\addauthor{Yunji Jung}{jungyj0105@kaist.ac.kr}{1}
\addauthor{Seokju Lee}{dltjrwn0322@kaist.ac.kr}{1}
\addauthor{Tair Djanibekov}{tair\_djan@kaist.ac.kr}{1}
\addauthor{Hyunjung Shim}{kateshim@kaist.ac.kr}{1}
\addauthor{Jong Chul Ye}{jong.ye@kaist.ac.kr}{1}
\addinstitution{
Korea Advanced Institute of Science
and Technology (KAIST)\\
Daejeon, Republic of Korea
}

\runninghead{Student, Prof, Collaborator}{BMVC Author Guidelines}

\begin{document}

\maketitle

\begin{abstract}
Text-guided non-rigid editing involves complex edits for input images, such as changing motion or compositions within their surroundings. Since it requires manipulating the input structure, existing methods often struggle with preserving object identity and background, particularly when combined with Stable Diffusion. In this work, we propose a training-free approach for non-rigid editing with Stable Diffusion, aimed at improving the identity preservation quality without compromising editability. Our approach comprises three stages: text optimization, latent inversion, and timestep-aware text injection sampling. Inspired by the success of Imagic, we employ their text optimization for smooth editing. Then, we introduce latent inversion to preserve the input image's identity without additional model fine-tuning. To fully utilize the input reconstruction ability of latent inversion, we suggest timestep-aware text injection sampling. This effectively retains the structure of the input image by injecting the source text prompt in early sampling steps and then transitioning to the target prompt in subsequent sampling steps. This strategic approach seamlessly harmonizes with text optimization, facilitating complex non-rigid edits to the input without losing the original identity. We demonstrate the effectiveness of our method in terms of identity preservation, editability, and aesthetic quality through extensive experiments.
\end{abstract}

%-------------------------------------------------------------------------
\section{Introduction}
%Diffusion models have become a new standard in image generation and have had a significant impact on text-to-image generation tasks. Text-guided diffusion models take text prompts as input and generate images that align well with the input text prompt. Building upon the success of text-guided diffusion models, there has been active research in text-guided image editing. This encompasses global style transfer~\cite{tumanyan2023plug,valevski2023unitune,wang2023stylediffusion,zhang2022inversion} and local editing that require changing object attributes (e.g., color, shape), image translation and adding details~\cite{hertz2022prompt,couairon2022diffedit,avrahami2022blended,brack2023ledits++}.

% \begin{figure*}[!t]
% \begin{center}
% \includegraphics[width=1.0\linewidth]{IJCAI2024/Images/first_page.pdf}
% \end{center}
% \vspace{-0.2in}
% \caption{Our method excels detailed non-rigid edits, such as changing the motion or composition of an object on real images directed by text prompts via Stable Diffusion. This figure showcases original images with their corresponding text-guided modification. Our training-free method preserves the original identity and context of the input images while achieving high-level conceptual alterations that seamlessly blend with the images' inherent characteristics.}
% \label{fig:teaser}
% \vspace{-0.2in}
% \end{figure*}

Among various image editing tasks, non-rigid editing involves changing the posture or composition of an object while preserving the background and object identity of the original image. It is useful for various applications, such as drawing comic books or videos, where gradual transitions in object actions or compositions over the input image are essential. This task is particularly more challenging than rigid editing, such as global style transfer and image translation, which focus on changing color or textures while retaining the image structure. 
%Instead, non-rigid editing requires modifying an object structure partially to align with the text prompt while maintaining its identity.

The challenges of non-rigid editing become more severe when combined with Stable Diffusion~\cite{rombach2022high}, primarily related to preserving image identity. Prior works in image editing address this problem through various approaches, including injecting attention maps of the source image into the target image generation process~\cite{hertz2022prompt,tumanyan2023plug,xiao2023fastcomposer,cao2023masactrl,wang2024unified,qiao2024baret}, model finetuning~\cite{ruiz2023dreambooth,valevski2023unitune,kawar2023imagic}, and target text optimization~\cite{gal2022image,han2023highly,kawar2023imagic}. However, the attention mechanism often leads to unnatural images as it forcefully combines the structure of the source image with the target, resulting in distorted objects and compositions. Also, model finetuning on an input image can lead to a severe overfitting problem and color distortion, especially when applied to Stable Diffusion.

To this end, our work explores improving identity preservation for non-rigid editing with Stable Diffusion by incorporating a text optimization framework. We focus upon the following key insight: text optimization on an input image has shown promise in enabling smooth changes of an object toward the desired editing concept. However, this hinges on the condition that the model faithfully reproduces the input image and ensures that small changes in text embedding do not significantly alter the output image’s resemblance to the input image~\cite{kawar2023imagic}.

Therefore, Imagic~\cite{kawar2023imagic} additionally introduced model finetuning, ensuring the model specialized in reproducing the input image. Nevertheless, as this method causes overfitting and color distortion problems, we make a pivotal change by replacing the model fine-tuning with latent inversion. This change maintains the integrity of the original data distribution, preserving the editability of text optimization. Additionally, we exploit the reconstruction power of latent inversion further by inserting source text embedding exclusively during the initial sampling steps. Then, we transition to target text embedding for the subsequent sampling steps, namely timestep-aware text injection. This concept is inspired by the recent study that reveals the pivotal role of the initial sampling steps in determining the coarse structure and outlines of an image. By selectively injecting source and target prompts considering the role of sampling timesteps, we can explore the smooth transition between identity preservation and editability. This approach allows us to find a sweet spot between these two factors.

Our simple yet effective approach effectively harmonizes with text optimization, achieving complex non-rigid editing without losing the original identity. 
%It does not require additional inputs other than one input image and text prompts. 
Unlike Imagic, our training-free method offers the advantage of reusing a pre-trained diffusion model regardless of images. Also, our method generates more natural and aesthetically pleasing images than attention-based methods.
We compare our edited results both quantitatively and qualitatively with other editing methods, showing superior editing performance on the Stable Diffusion in terms of identity preservation, editability, and aesthetic quality. Moreover, we prove each component of our method is effective through ablation experiments and show the applicability of our method on the different image domain such as Anything-v4.

\section{Related Works}
\subsection{Text-guided Image Manipulation}
Text-guided image manipulation involves editing an image based on a given text prompt while preserving the original input image identity. In the realm of this technique, the predominant approach for preserving the original image identity involves the use of an attention mechanism. This mechanism integrates self or cross-attention maps from the input image into the generation process of an edited image~\cite{hertz2022prompt,tumanyan2023plug,cao2023masactrl,parmar2023zero,wang2024unified,qiao2024baret}. Additionally, other methods blend noise maps of the source image and target image, thereby maintaining the background from the source image~\cite{couairon2022diffedit,avrahami2022blended,huang2023region}.
Other directions include model finetuning~\cite{ruiz2023dreambooth,valevski2023unitune,kawar2023imagic,li2023layerdiffusion} or text optimization~\cite{gal2022image,han2023highly,kawar2023imagic}, which focus on adapting a model or target text to closely match the input image, even to the extent of overfitting. Furthermore, score distillation has been introduced for image editing, where the input image identity is continuously distilled into the target image during the editing process~\cite{hertz2023delta}. 
Finally, several approaches~\cite{brack2024sega,tsaban2023ledits,brack2023ledits++} utilize semantic guidance by adding an edit term in the original classifier-free guidance formula~\cite{ho2021classifier}.

\subsection{Diffusion Inversion}
For real image editing, the concept of inversion has been widely used to uncover the latent representation that faithfully reconstructs the input real image. DDIM inversion~\cite{song2020denoising} excels at reconstructing the original image for unconditional image generation scenarios. However, it faces challenges in conditional cases where the use of classifier-free guidance (CFG)~\cite{ho2021classifier} magnifies the errors, resulting in decreased editability and reconstruction quality. This drawback of DDIM gave rise to active research in inversion specialized for conditional image generation. 
Null-text inversion~\cite{mokady2023null} improves the inversion quality by optimizing null text embedding $\emptyset$ at each diffusion timestep. 
On the other hand, Direct Inversion~\cite{ju2023direct} takes a different approach by disentangling source and target diffusion branches, eliminating optimization. This strategy boosts content preservation quality and edit fidelity within a short time.
EDICT~\cite{wallace2023edict} offers a mathematically exact inversion solution through an alternative inversion process that involves a pair of coupled noise vectors.
Edit-Friendly inversion~\cite{huberman2023edit} strongly imprints the input image onto latent noise maps by making them correlated across timesteps, enabling fast and effective editing without the need for optimization.
%We utilize DDIM, Null-text, and Direct inversion for our experiments as they are the representative methods.
%More explanations on classifier-free guidance and inversion methods that we utilized are in Appendix.

\section{Preliminary}
\label{sec:analysis}
\noindent\textbf{Stable Diffusion Model.} 
%Diffusion models generate an image $z_0 \sim q(z_0)$ by iteratively denoising a noise distribution. A forward process samples $x_t$ at timestep $t \in [1,...,T]$ by adding a Gaussian noise $\beta_t \in (0, 1)$ at $z_0$ using Equation~\ref{eq_ddpm_forward}:
%\begin{equation} \small %\label{eq_ddpm_forward}
%x_t = \sqrt{\bar{\alpha}_t}z_0 + \sqrt{1-\bar{\alpha}_t}\epsilon
%\end{equation}
%where $\epsilon \sim \mathcal{N}(0,\boldsymbol{I})$, $\alpha_t = 1 - \beta_t$, and $\bar{\alpha}_t = \prod_{i=0}^{t}\alpha_i$. 
%A neural network $\epsilon_{\theta}$ is then trained to predict $\epsilon$ of $x_t$ using the objective:
%\begin{equation} \small 
%L_{\text{simple}} = E_{t,z_0,\epsilon}\left[ || \epsilon - \epsilon_{\theta}(x_t, t) ||^2 \right]
%\end{equation}
%\noindent where $t \sim Uniform[1,T]$.
%During the reverse process, $x_{t-1}$ is obtained by denoising $x_t$ using Equation~\ref{eq_ddpm_reverse}: 
%\begin{equation} \small 
% \label{eq_ddpm_reverse}
%x_{t-1} 
%= \frac{1}{\sqrt{1-\beta_t}}
%\left(x_t - \frac{\beta_t}{\sqrt{1-\overline{\alpha}_t}}\epsilon_{\theta}(x_t,t)\right)
%+\sigma_{t}\epsilon.
%\end{equation}
%where $\sigma_t := \frac{1 - \bar\alpha_{t-1}}{1 - \bar\alpha_t}\beta_t$.
%This process is performed iteratively starting from $x_T \sim \mathcal{N}(0,\boldsymbol{I})$ until we obtain a clean image $z_0$. 
Diffusion models aim to estimate a data distribution via a latent variable model. They perform a forward process that gradually adds a Gaussian noise on a clean image at each timestep $t \in [1,..., T]$ until it reaches a noise distribution. Conversely, the reverse process inversely denoises the noise distribution in order to generate a final clean image.
The Stable Diffusion model~\cite{rombach2022high}, an open-source model, conducts these processes on the latent space. Specifically, given an image $\boldsymbol{x} \in \mathbb{R}^{H \times W \times 3}$, it encodes $\boldsymbol{x}$ into a latent representation $\boldsymbol{z}_0 = \mathcal{E}(\boldsymbol{x})$ using a VQ-GAN~\cite{esser2021taming} encoder $\mathcal{E}$. The model is trained on this latent space, applying the forward process to $\boldsymbol{z}_0$. Likewise, $\boldsymbol{z}_0$ is obtained from the latent variable $\boldsymbol{z}_T$ using the reverse process. Finally, the generated image is computed by decoding $\boldsymbol{z}_0$ back to the image space ($\boldsymbol{x} = \mathcal{D}(\boldsymbol{z}_0)$). Also, an important function of Stable Diffusion is its capability to facilitate text-guided image generation. This is achieved by inputting a text embedding $\tau_\phi(\boldsymbol{y})$, which is a text prompt $\boldsymbol{y}$ processed by CLIP~\cite{radford2021learning} text encoder $\tau_\phi$. The text-guided conditional model is trained using the following objective:
\begin{equation} \small 
\mathcal{L}_{\text{cond}} = \mathbb{E}_{t,\boldsymbol{z}_0,\boldsymbol{\epsilon}}\left[ || \boldsymbol{\epsilon} - f_{\theta}(\boldsymbol{z}_t, t, \tau_\phi(\boldsymbol{y})) ||^2 \right],
\end{equation}
where $f_{\theta}$ is a neural network with parameter $\theta$, $t \sim \texttt{Uniform}[1,T]$, $\boldsymbol{\epsilon} \sim \mathcal{N}(0,\boldsymbol{I})$, and $\boldsymbol{z}_t$ is a noise added version of $\boldsymbol{z}_0$ using the forward process.
Our goal is to improve non-rigid editing with Stable Diffusion, aiming to enhance the accessibility of a publicly available model. 
%Detailed explanations of the forward and reverse process of diffusion models are available in Appendix.

%Non-rigid editing demands two properties: 1) identity preservation, where the edited image should maintain the object identity and background of the original input image, and 2) text-guided generation quality, in which the edited result should align seamlessly with the target text. 

\vspace{2mm}
%\noindent\textbf{Challenges in Non-rigid Editing.} Compared to rigid editing, such as global style transfer, image translation, and appearance transformation, non-rigid editing stands out as highly challenging. Specifically, rigid editing handles relatively simple color adjustments while maintaining the image structure intact, which model filters can easily capture. Conversely, non-rigid editing involves understanding and partially modifying an underlying structure of the original object while maintaining its identity. For that, the model should faithfully reconstruct the original input while selectively editing the structural components that align with the target prompt. Since faithful reconstruction has a trade-off relationship with diverse editing in most text-guided image generation models, non-rigid editing often lacks identity preservation. Therefore, improving the identity preservation ability of the model is the main issue in enhancing the non-rigid editing performance.  

\begin{figure*}[!t]
\begin{center}
\includegraphics[width=1.0\linewidth]{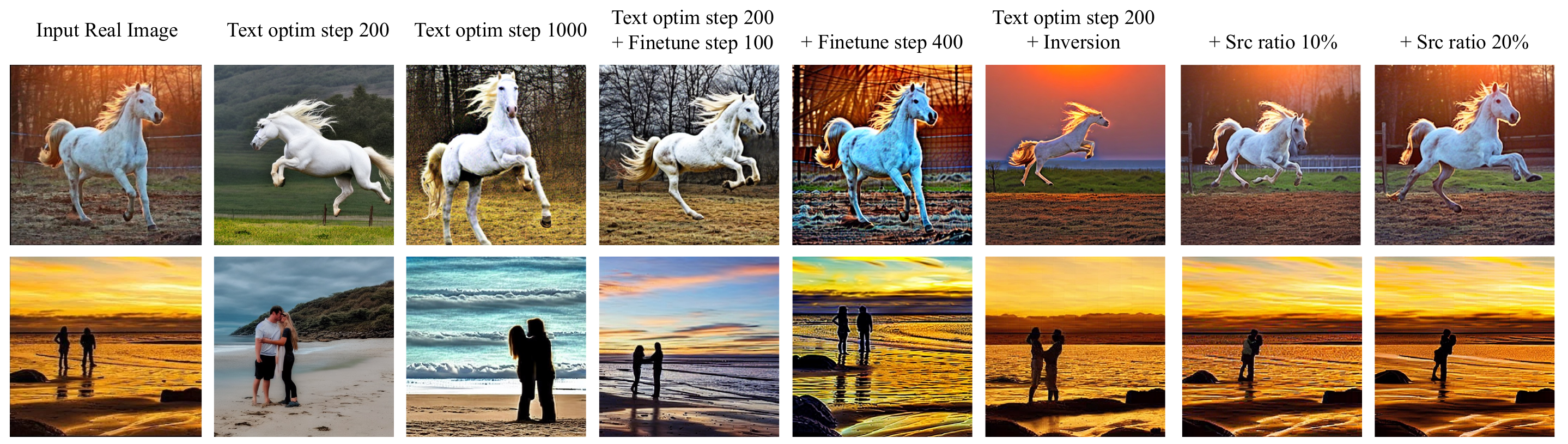}
\end{center}
\vspace{-0.18in}
\caption{Ablation on Imagic and Our method. The images are edited according to the target text: ``A photo of a white horse jumping'' and ``A photo of a couple hugging on a beach''. 
%Text optimization minimally affects identity preservation. Model fine-tuning from Imagic enhances image resemblance to the input, often at the expense of color fidelity. Conversely, our latent inversion stabilizes the overfit tradeoff and avoids color distortion. To further improve identity preservation, we incorporate source text embedding during the initial sampling timesteps.
}
\label{fig:ablation}
\vspace{-0.2in}
\end{figure*}

\vspace{2mm}
\noindent\textbf{Imagic for Non-rigid Editing.} Imagic~\cite{kawar2023imagic}, a pioneering method for non-rigid editing, introduced a two-step process: text optimization and model fine-tuning. 
Firstly, they optimize the target text embedding to best match the input image. Specifically, the target text is encoded to a text embedding $\boldsymbol{e}_{tgt} \in \mathbb{R}^{N \times d}$ via a text encoder~\cite{raffel2020exploring}, where $N$ is the number of tokens in the target text and $d$ is the token embedding dimension. Following this, they freeze parameters
of the pre-trained diffusion model $f_\theta$ and optimize the target text embedding $\boldsymbol{e}_{tgt}$ using the diffusion model objective:
\begin{equation} \label{eq_objective} \small
\boldsymbol{e}_{opt}  =\arg \min \mathbb{E}_{t, \boldsymbol{z}_0, \boldsymbol{\epsilon}}\left[ \left\|\boldsymbol{\epsilon}- f_{\theta}\left(\boldsymbol{z}_t, t, \boldsymbol{e}_{tgt})\right)\right\|^{2}\right]. 
\end{equation}
Though the target text embedding is optimized to be matched with the input image, $\boldsymbol{e}_{opt}$ often does not show a perfect representation of the input image. Additional optimization does not improve identity preservation but causes a runtime overhead problem (\emph{Text optim step 200}, \emph{Text optim step 1000} in Figure~\ref{fig:ablation}). Therefore, Imagic uses model fine-tuning as a firm identity preservation strategy to make the model overfitted to the input image. 
It freezes the optimized embedding and then fine-tunes the model parameters $\theta$ using the same loss function in Equation \ref{eq_objective}.

Finally, during the sampling phase, Imagic linearly interpolates the original target text embedding with the optimized one using Equation \ref{eq_interpolate} and passes this through the model to generate an edited image. Here, it can control the degree of editing by varying an interpolation ratio $\alpha$. This aims to establish a smooth transition of text embedding toward the desired editing concept while maintaining the original identity. 
\begin{equation} \label{eq_interpolate}
\boldsymbol{e}_{int} = \alpha \cdot \boldsymbol{e}_{tgt} + (1 - \alpha) \cdot \boldsymbol{e}_{opt}
\end{equation}
It is worth noting that Imagic is particularly effective when applied to Imagen~\cite{saharia2022photorealistic}, a large-scale generative model only accessible for business. It successfully applies the editing concept while faithfully maintaining the original identity and background of the original image. However, the same recipe encounters limitations when applied to Stable Diffusion, as mentioned in previous works~\cite{han2023highly,brack2023ledits++}. Our experiments also affirm the unsatisfactory editing performance of Imagic on Stable Diffusion (Sec~\ref{sec:results}), with a noticeable deficiency in identity preservation.
%As a result, we aim to nurture the strength of text optimization also on Stable Diffusion and enhance the non-rigid editing performance.

\begin{figure*}[!t]
\begin{center}
\includegraphics[width=1.0\linewidth]{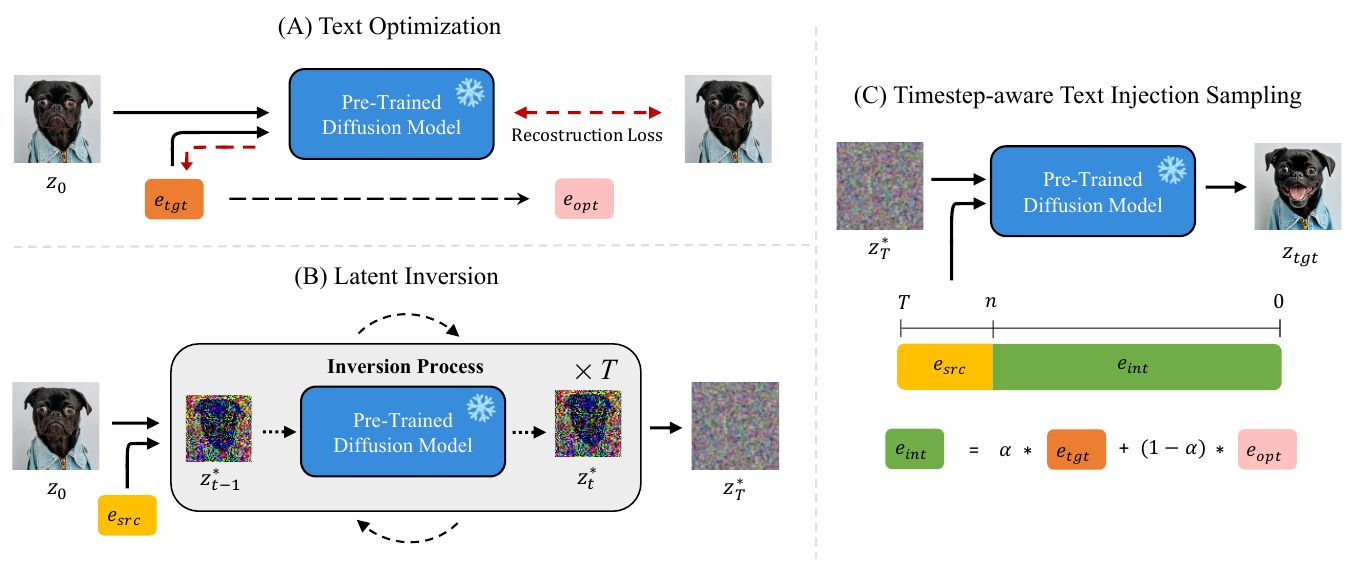}
\end{center}
\vspace{-0.2in}
\caption{Overview of our method.
% : (A) text optimization, (B) latent inversion, and (C) timestep-aware injection sampling. 
% (A) The target text is encoded into an optimized embedding to match the input image. (B) Null-text inversion is employed for additional identity preservation. (C) The source text embedding $\boldsymbol{e}_{src}$ instead of an interpolated embedding $\boldsymbol{e}_{int}$ is injected during the initial sampling steps to anchor the image's global structure more effectively. 
}
\label{fig:method}
\vspace{-0.2in}
\end{figure*}

\section{Method}
\label{sec:method}

In this section, we identify the problem of Imagic combined with Stable Diffusion and introduce our key contributions to enhancing non-rigid editing performance. An overview of our method is demonstrated in Figure~\ref{fig:method}.

We follow the text optimization from Imagic to preserve the original identity and achieve smooth non-rigid editing (See Figure~\ref{fig:method} (A)). As text optimization alone is insufficient for a perfect match with the input image, Imagic solves this problem via model finetuning. However, in the Stable Diffusion setting, we found that the model finetuning presents several significant drawbacks that adversely affect editing quality. Firstly, it triggers a severe tradeoff between (1) \emph{overfitting}, where the model overly memorizes the original image and thus cannot faithfully incorporate the text prompt well, or (2) \emph{the lack of identity preservation}, where the edited image aligns with the target text but it fails to preserve the original image's identity. Figure~\ref{fig:ablation} provides visual evidence that the edited image closely resembles the input image as we finetune the model longer. Conversely, minimal finetuning produces an arbitrary image that bears little resemblance to the input image, indicating the lack of identity preservation (\emph{Text optim step 200 + Finetune step 100}, \emph{+Finetune step 400} in Figure~\ref{fig:ablation}). Model finetuning thus hardly satisfies both identity preservation and editability simultaneously. (More ablation experiments on finetuning steps are in Appendix.) Even worse, it causes color distortion, disrupting the data distribution. This is because model training aims to reach the data distribution of the training dataset, which in this context consists of only a single input image.

Instead of model fine-tuning, we utilize latent inversion to maintain the original identity (See Figure~\ref{fig:method} (B)). The inversion process is defined by Equation~\ref{eq:inversion}:
\begin{align} \small \label{eq:inversion} 
    \bm{z}^*_{t+1} = & \sqrt{\frac{\alpha_{t+1}}{\alpha_t}}\bm{z}^*_t + \nonumber\\
    & \sqrt{\alpha_{t+1}} \left( \sqrt{\frac{1}{\alpha_{t+1}} - 1} - \sqrt{\frac{1}{\alpha_t} - 1} \right) f_{\theta}(\bm{z}^*_t, t, \boldsymbol{e}_{src}),
\end{align} 
where $\alpha_t$ is a noise schedule at timestep $t$ and $\boldsymbol{e}_{src}$ is a source text embedding. As the inversion produces a latent that faithfully reconstructs the original image, the use of inverted latent for sampling can effectively maintain the original identity with text-guided editing. Moreover, the inversion method no longer suffers from (1) the tradeoff between overfitting and editability and (2) color distortion.

While employing the inverted noise during sampling process helps maintain the overall color and context of the original image, introducing a new text prompt different from the source text can cause deviations in image configuration, such as the size and position of the object, compared to the original image. To address this issue, we leverage a well-known observation that diffusion models generate global structures at the initial timesteps and fine details toward the end~\cite{choi2022perception}. Specifically, we introduce an additional regularization depending on sampling timesteps (See Figure~\ref{fig:method} (C)). During the initial steps ($n \rightarrow T$), we incorporate the source text embedding and then transition to interpolated embeddings for the subsequent timesteps: 
\begin{equation} \small 
\boldsymbol{e}_{input} = 
\begin{cases*}
\boldsymbol{e}_{src},       & if  n $\le$ t $\le$ T\\
\boldsymbol{e}_{int},    & otherwise
\end{cases*}
\end{equation}
This strategy helps to closely adhere to the global structure of the original image, a key component of the image identity, during the early stage of the editing process. Since then, the editing concept is applied to the subsequent sampling steps, effectively incorporating editing concepts without altering the image identity. We show its substantial impact on preserving the identity through ablation studies in Section~\ref{section:ablation}.

\section{Experiments}
\label{sec:results}
\subsection{Implementation Details}
% \noindent\textbf{Dataset.}
% We employ the Textual Editing Benchmark (TEdBench), introduced by Imagic~\cite{kawar2023imagic}. It comprises input images with target texts describing desired complex non-rigid editing.

\vspace{2mm}
\noindent\textbf{Baselines.}
For a comprehensive comparison, we evaluate our method against the following competitive techniques: Imagic~\cite{kawar2023imagic}, Null-text inversion~\cite{mokady2023null}, DDS~\cite{hertz2023delta}, Masactrl~\cite{cao2023masactrl} and Edit-Friendly~\cite{huberman2023edit}. Notably, Imagic and Masactrl are specialized for the non-rigid editing task. Null-text inversion and Edit-Friendly represent prominent methods designed for natural image editing via inversion. DDS proposes a new paradigm of image editing using score distillation. In our evaluations, we run the released codes of all baseline models using pre-trained Stable Diffusion v1.4 with publicly available checkpoints. Unfortunately, we could not include other non-rigid editing methods, such as LayerDiffusion~\cite{li2023layerdiffusion} and BARET~\cite{qiao2024baret}, as their codes are not released as of our submission date. Details on experiments are in Appendix.
% For sampling, we use a conventional setting of DDIM with 50 timesteps ~\cite{cao2023masactrl,huang2023kv,li2023layerdiffusion} for all methods. For our method using DDIM inversion~\cite{song2020denoising}, we used a guidance scale of 1 for both inversion and sampling stages, as it is known that using a larger guidance scale leads to suboptimal results. For our method using Null-text inversion~\cite{mokady2023null}, we used a guidance scale of 7.5 for both inversion and sampling following previous works~\cite{cao2023masactrl,huang2023kv}.  

\vspace{2mm}
\noindent\textbf{Evaluation Metrics.}
We utilize CLIP-T score~\cite{radford2021learning} for measuring target text alignment and LPIPS~\cite{zhang2018unreasonable}, PSNR, and MSE for measuring input image identity preservation quality of edited images. However, these metrics are incapable of evaluating the aesthetic quality of the edited image; they cannot penalize the distorted or unnatural outputs. To mimic the standard of human visual inspection, we additionally employ Aesthetic score~\cite{aesthetic_classifier}, Pickscore~\cite{kirstain2024pick}, and Imagereward~\cite{xu2024imagereward} which measure how much people like the image on average.
%CLIP score concerning the target text measures the cosine similarity between the embeddings of the target text prompt and the edited image. LPIPS score measures the difference between the original input image and the edited image in the VGG~\cite{simonyan2014very} feature space. However, these two metrics are incapable of evaluating the aesthetic quality of the edited image; they cannot penalize the distorted or unnatural outputs. To mimic the standard of human visual inspection, we additionally employ an Aesthetic score~\cite{aesthetic_classifier}, which measures how much people like the image on average. Few MLP layers are added on the CLIP ViT-L/14, and only the MLP layers are finetuned to predict the aesthetic score using MSE or MAE loss. AVA~\cite{murray@AVA} and Simulacra~\cite{John@simulacra} are used for training the predictor, which are datasets that collect human ratings of 1 to 10 on real and generated images each. This metric is used to clean the LAION-5B~\cite{schuhmann2022laion} dataset.

\begin{figure*}[!t]
\begin{center}
\includegraphics[width=1.0\linewidth]{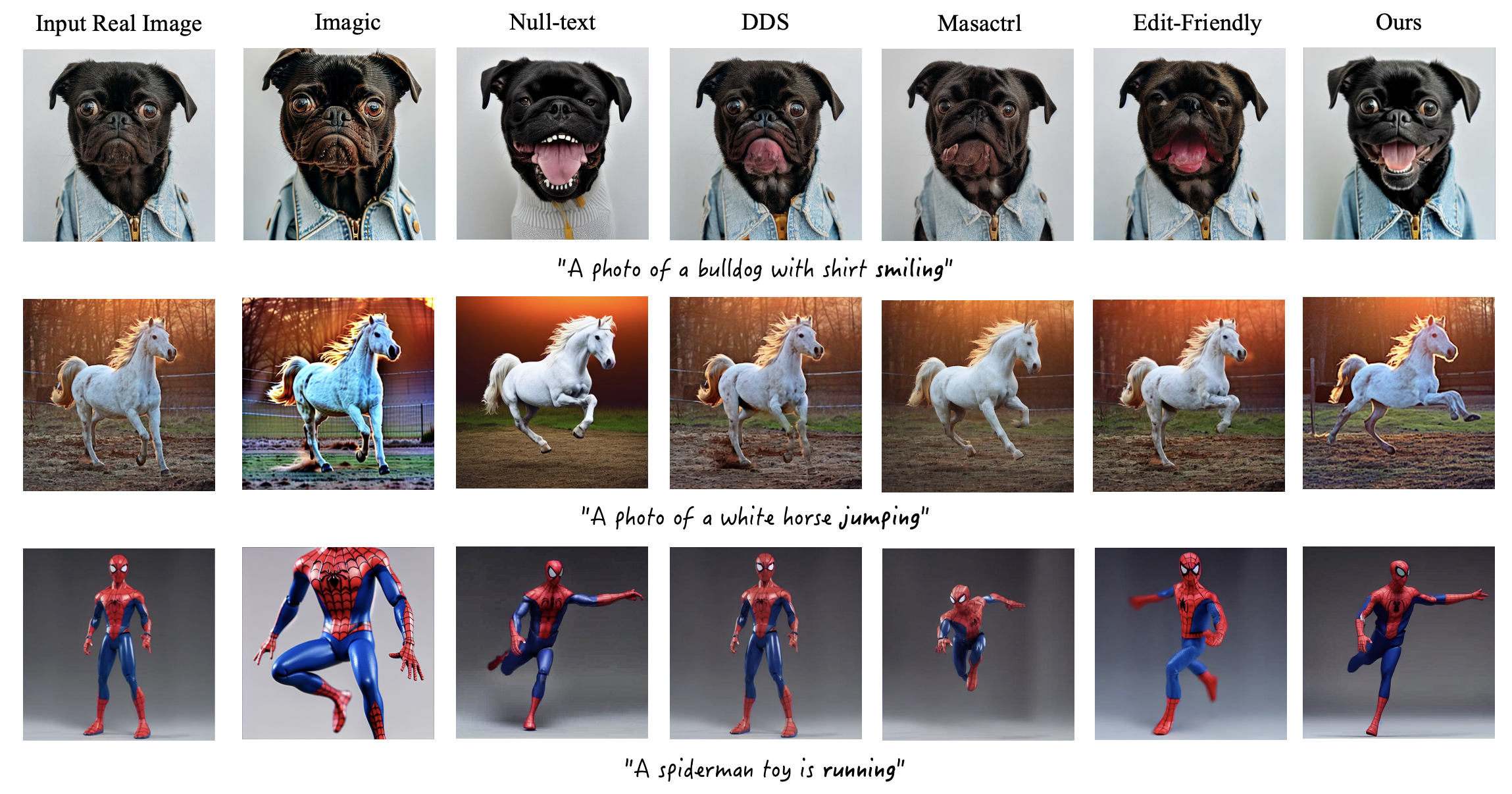}
\end{center}
\vspace{-0.2in}
\caption{Qualitative comparison with state-of-the-art methods on Stable Diffusion v1.4. 
%While other methods suffer from either overfitting the input image or lacking identity preservation, our method shows the best structure preservation and edit fidelity.
}
\label{fig:qualitative}
\vspace{-0.2in}
\end{figure*}

\subsection{Qualitative Evaluation}
We compare the editing results of our method to state-of-the-art methods using a pre-trained Stable Diffusion v1.4 in Figure~\ref{fig:qualitative}. We show our results using Null-text inversion as it provides the best editing performance (Figure~\ref{fig:inversion_type}). For a fair comparison, we also applied Null-text inversion in Masactrl, which originally used DDIM inversion. It results in improved identity preservation and editability of Masactrl. We tried various hyperparameters of each method and chose the best result for each.

Firstly, Null-text inversion alone for editing, in which we just input an inverted latent and a target text through the model, shows low identity preservation. It does not retain the original structure of the input image and simply pursues generating an image that aligns well with the text prompt. This is expected because there is no other constraint for identity preservation during the sampling process. 
Meanwhile, Imagic has a serious tradeoff between overfitting and editability. Model finetuning especially causes a severe overfitting problem, which distorts the data distribution, whereas a random noise without finetuning during the sampling process ruins the structure preservation. Despite the extensive hyperparameter search, it is hard to find a sweet spot between these two factors (See Appendix). Besides, DDS also suffers from overfitting to the input image, especially when handling non-rigid editing. This shows that score distillation has trouble changing geometry.
On the other hand, Edit-Friendly and Masactrl attempt to apply an editing concept while maintaining the original identity by improving latent noise space and utilizing attention maps, respectively. When an image has no complicated background and contains a single object, such as a Spiderman image in the third row of Figure~\ref{fig:qualitative}, their editing quality is sufficiently high. However, they show unnatural edited results in other cases. In the first row of Figure~\ref{fig:qualitative}, a bulldog opens its mouth but does not smile in Edit-Friendly, and the bulldog shows a grotesque mouth in Masactrl. In the second row, a horse jumps in both methods, but the hind legs of the horse remain similar to the input image, resulting in an unnatural jumping appearance. 
%This is attributed to the difference in object appearance and poses between the source image and target image when naively generated using the original source text and target text. Example images are shown in Appendix.???. As the appearance gap between the target image and the source image intensifies, inserting the attention maps of the source image into the target image sampling process often results in abnormal objects with unnatural appearances and poses. 
Overall, our method clearly shows the editing concept while faithfully preserving the original identity, like a bulldog with a big smile and a horse jumping dynamically. These visual trends and the superiority of our method consistently hold in other results. More qualitative results are demonstrated in Appendix, including results using the pretrained Anything-v4 model.

% \vspace{2mm}
% \noindent\textbf{Results with Anything-V4.} We also perform our method on a pretrained anime-style model, Anything-V4. We generated an image using the model and then performed editing just like the real image editing setting.  Figure~\ref{fig:Anything} shows various editing results using diverse text prompts. We can change the posture, facial expression, and viewpoint smoothly while maintaining the original structure of the input image. This proves that our method generalizes well regardless of the image domain.  More qualitative results are shown in Appendix.

% \begin{figure}[!t]
% \begin{center}
% \includegraphics[width=1.0\linewidth]{IJCAI2024/Images/Anythin-v4_final_2.pdf}
% \end{center}
% \vspace{-0.2in}
% \caption{\textbf{Anything-v4.} This figure showcases the adaptability of our model on an Anything-V4 anime-style model. Our method effectively edits the input image's posture and viewpoint with various text prompts, smoothly preserving the original structure. }
% \label{fig:Anything}
% \vspace{-0.2in}
% \end{figure}

\subsection{Quantitative Evaluation}
\begin{figure}[!t]
\begin{center}
\includegraphics[width=1.0\linewidth]{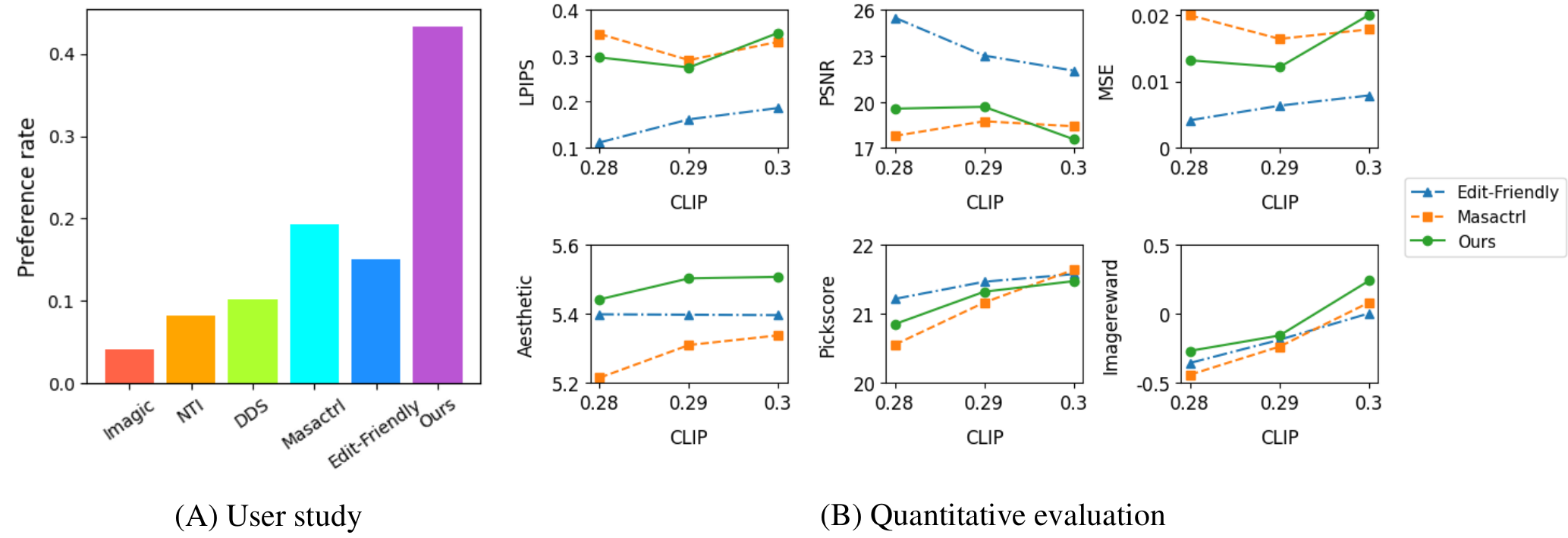}
\end{center}
\vspace{-0.2in}
\caption{User study and quantitative comparison.}
\label{fig:quantitative}
\vspace{-0.2in}
\end{figure}

Due to inherent inaccuracies and ambiguities in quantitative evaluation of non-rigid editing, we conduct a user study. We utilized the TedBench dataset~\cite{kawar2023imagic} to generate 50 edited images of each baseline model. A set of input real image, target text and 6 edited images using each model is shown to the user. The user is asked to choose one among edited images that best satisfies both two conditions simultaneously: The edited image 1) maintains background and appearance of the object in the original input image and 2) aligns with the target text. We randomly shuffled the order of edited images so that the user doesn't know a model used to generate each image. We accumulated 38 respondents, resulting in total 1900 (38 $\times$ 50) responses. Figure~\ref{fig:quantitative} (A) shows the preference rate of each method, which represents the proportion of times the corresponding method was selected out of 1900 responses. Ours receives the highest preference rate, 0.43, which is more than twice as high as the preference rate of the second-ranked Masactrl (0.19). It indicates that our method surpasses other models in terms of both similarity to the input image and text alignment. 

We further conduct a quantitative evaluation on the models that ranked within the top three in user study: Edit-Friendly, Masactrl, and Ours. 
%The experiment is carried out using the pre-trained Stable Diffusion v1.4. To ensure a fair comparison, we use Null-text inversion for both ours and Masactrl. 
We utilized the same target text prompts used in user study to generate 50 edited images, producing five variants for each edited image with different seeds, resulting in a total of 250 images. In Figure~\ref{fig:quantitative} (B), we present various metrics of each method at different CLIP-T scores, where CLIP-T changes depending on hyperparameters. 
%LPIPS ($\downarrow$), PSNR ($\uparrow$), MSE ($\downarrow$) show similarity to the original input image while Aesthetic score ($\uparrow$), Pickscore ($\uparrow$) and Imagereward ($\uparrow$) indicate aesthetic assessment and human preference.
Generally, Edit-Friendly receives the best score in LPIPS ($\downarrow$), PSNR ($\uparrow$) and MSE ($\downarrow$) showing that it has a strong tendency to preserve the identity of the original image. Compared to Edit-Friendly, Ours and Masactrl edit images in a way that align more closely with the text prompt. For non-rigid editing which involves changes in motion of objects, big changes in the structure of objects should be allowed. Therefore, strictly preserving the structure of the original image disturbs editing. Our user study proves that our method well balances input image similarity and editability. This success can be attributed to our inversion and timestep-aware text injection sampling, which preserve the original identity without disrupting the data distribution. This enables text optimization to smoothly transition towards the desired editing concept while maintaining the background and object identity faithfully. Also, our method receives the highest score in Aesthetic score ($\uparrow$) and Imagereward ($\uparrow$) indicating that our edited images are more preferred in terms of human preference.
% Imagic consistently yields the lowest CLIP score across all settings, indicating that its editability stays far behind ours and Masactrl. Our method exhibits the highest CLIP score in most settings. It implies that our method effectively preserves the input image's identity while accurately conveying the editing concept compared to other methods. Regarding the Aesthetic score, our method achieves the highest score in all the settings. It highlights that our method has both superior editability and aesthetic appeal compared to other methods. 
%Imagic tends to overfit the input image, maintaining the naturalness of the image structure, whereas color gets distorted. Consequently, its aesthetic score lags behind ours, but not the least. Finally, 
Masactrl, on the other hand, presents the lowest score in Aesthetic score, Pickscore and Imagereward in general, consistent with its tendency to generate unnatural images with distorted object appearances and poses. This can be attributed to the difference in object appearance and poses between the source and target images, each generated using the original source and target text, respectively. Example images are available in Appendix. As the appearance gap between the target image and the source image intensifies, inserting the attention maps of the source image into the target image sampling process often results in abnormal objects with unnatural appearances and poses. 
%Though there is a point where Masactrl has a higher CLIP score compared to ours, its low Aesthetic score shows that it has difficulty generating an edited image in which the editing concept seamlessly blends with the object's intrinsic characteristics.

% \begin{figure}[!t]
% \begin{center}
% \includegraphics[width=1.0\linewidth]{IJCAI2024/Images/quantitative_3.png}
% \end{center}
% \vspace{-0.2in}
% \caption{\textbf{Quantitative results.} The graph represents (A) CLIP scores ($\uparrow$), indicating alignment with target text, and (B) Aesthetic scores ($\uparrow$), under various image quality levels as LPIPS scores change. Our method outperforms Imagic and Masactrl in both CLIP and Aesthetic scores in most settings, demonstrating superior editability and aesthetic quality. This highlights the effectiveness of our method, latent inversion with timestep-aware text injection sampling, in preserving the original identity and data distribution, leading to more natural and concept-aligned edits.}
% \label{fig:quantitative}
% \vspace{-0.2in}
% \end{figure}

\subsection{Ablation Experiments}
\label{section:ablation}
%We show the effectiveness of each component of our method through ablation experiments on inversion type and source text injection ratio. 

\begin{figure}[t]
\begin{center}
\includegraphics[width=1.0\linewidth]{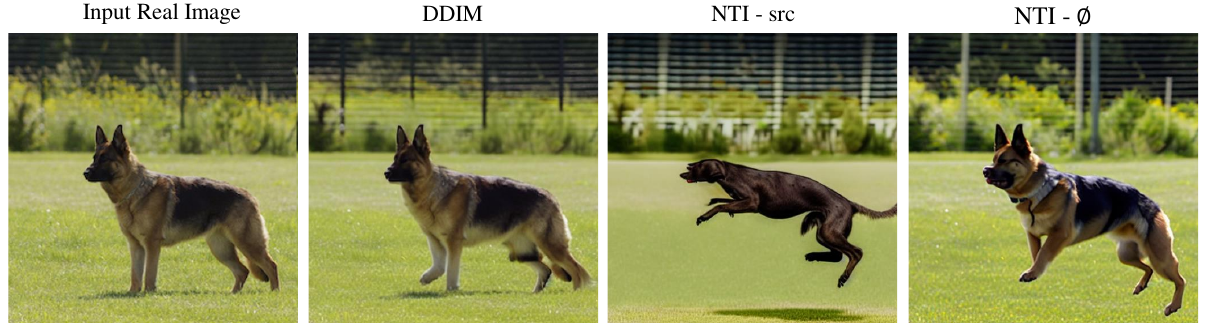}
\end{center}
\vspace{-0.2in}
\caption{Ablation on inversion type.
%We demonstrate our editing results with three inversion types: DDIM, Null-text inversion with source (\emph{NTI-src}), and without source embedding (\emph{NTI-w/o src}). 
%The results confirm that DDIM is less favorable to the editing task than NTI, where setting null text instead of source text often produces more plausible results.
}
\label{fig:inversion_type}
\vspace{-0.2in}
\end{figure}

\vspace{2mm}
\noindent\textbf{Inversion Type.}
Figure~\ref{fig:inversion_type} demonstrates the impact of different inversion methods: DDIM~\cite{song2020denoising} and Null-text inversion~\cite{mokady2023null}. Each image is edited with the target text ``A photo of a jumping dog.'' When employing DDIM inversion for our method, it is highly close to the input image but has low editability. This again proves that DDIM inversion is suboptimal for real image editing applications. On the other hand, Null-text inversion is known to improve reconstruction and editing quality, particularly when using classifier-free guidance with a guidance scale greater than 1, as compared to DDIM inversion. Interestingly, when inputting a source prompt (e.g. ``A photo of a dog") following the original method, the preservation quality sometimes significantly deteriorates while focusing on editing (\emph{NTI - src}). Conversely, when we set a source text prompt equal to null text (`'), the edited images show significant improvement in structural preservation (\emph{NTI - $\emptyset$}). Therefore, when the reconstruction performance falls behind, employing the null text for the source prompt can enhance the overall reconstruction ability. Overall, Null-text inversion demonstrates superior performance in editing quality to DDIM.

%\noindent\textbf{Text Optimization Steps.}
%We show the effect of text optimization steps in Figure.~\ref{fig:finetune_optim}. Each image is the edited results with a target text ”A photo of a white horse jumping”. As we optimize the target text longer, the edited image contains more information about the input image so the preservation quality of the edited images improves. If we optimize it too long, it totally gets overfitted to the input image which is improper for editing. The optimization step between 200 ~ 300 usually shows the best result.

\vspace{2mm}
\noindent\textbf{Source Text Injection Ratio.}
Our timestep-aware sampling strategy injects the source text embedding at the beginning of the sampling process instead of an interpolated embedding to improve identity preservation. Figure~\ref{fig:ablation} includes the ablation study with different source text input ratios. Given that we use DDIM sampling with 50 steps, inserting the source text for the initial 10 steps corresponds to a 20\% source text input ratio. We present results under three different scenarios: a ratio of 0\%, 10\%, and 20\%, depicted as \emph{Text optim step 200 + Inversion}, \emph{+ Src ratio 10\%}, and \emph{+ Src ratio 20\%} in Figure~\ref{fig:ablation}, respectively. Notably, as the source text input ratio increases, the edited images increasingly retain the original structure of the input image. We empirically observe that inserting the source text about 5 to 20 steps, which corresponds to a ratio of 10\% to 40\%, leads to the most favorable editing results depending on the degree of desired structural changes.

\section{Conclusion}
Our work has advanced the capabilities of non-rigid editing with Stable Diffusion, particularly focusing on identity preservation. 
Our method comprises text optimization, latent inversion, and timestep-aware text injection sampling. Based on the analysis of the previous study, we chose text optimization as the baseline as it allows a smooth transition toward a desirable editing concept. To overcome the limitations of model fine-tuning from Imagic, we suggest utilizing the reconstruction power of latent inversion. To further improve the identity preservation quality, we introduce timestep-aware text injection sampling that selectively applies source and target prompts depending on sampling timesteps. 
Our method presents superior performance in both qualitative and quantitative comparisons with other editing methods on Stable Diffusion. Our method does not trigger overfitting tradeoffs and color distortion, unlike model finetuning in Imagic. Also, we can provide natural and aesthetically pleasing images compared to attention-based methods. Finally, we can extend the usage of our method for other image domains like Anything-V4.

\bibliography{main}
\end{document}